\documentclass{article}
\usepackage{spconf,amsmath,graphicx,caption,multirow, booktabs, float, afterpage, placeins, url}

\title{Understanding zero-shot Rare Word Recognition improvements through LLM Integration}
\name{Haoxuan Wang}
\address{ucabhw2@ucl.com}

\begin{document}
%
\maketitle
\begin{abstract}
In this study, we investigate the integration of a large language model (LLM) with an automatic speech recognition (ASR) system, specifically focusing on enhancing rare word recognition performance. Using a 190,000-hour dataset primarily sourced from YouTube, pre-processed with Whisper V3 pseudo-labeling, we demonstrate that the LLM-ASR architecture outperforms traditional Zipformer-Transducer models in the zero-shot rare word recognition task, after training on a large dataset. Our analysis reveals that the LLM contributes significantly to improvements in rare word error rate (R-WER), while the speech encoder primarily determines overall transcription performance (Orthographic Word Error Rate, O-WER, and Normalized Word Error Rate, N-WER). Through extensive ablation studies, we highlight the importance of adapter integration in aligning speech encoder outputs with the LLM's linguistic capabilities. Furthermore, we emphasize the critical role of high-quality labeled data in achieving optimal performance. These findings provide valuable insights into the synergy between LLM-based ASR architectures, paving the way for future advancements in large-scale LLM-based speech recognition systems.
\end{abstract}
\section{Introduction}
\label{sec:intro}
\footnote{This work was done during Haoxuan's tenure at Zoom Communication Inc. He has now left the company.}
Automatic Speech Recognition (ASR) systems have significantly advanced, achieving impressive accuracy in transcribing spoken language. However, accurately recognizing rare words, such as proper nouns, technical terms, and uncommon expressions, remains a persistent challenge. These rare words often have limited representation in training datasets, leading to higher error rates and impacting the overall performance of ASR systems~\cite{he2023edcec}.

Recent research has explored various strategies to address this issue. For example, post-processing methods that focus on error detection and context-aware correction have been proposed to enhance rare word recognition~\cite{he2023edcec}. Furthermore, meta-learning approaches have been investigated to enable few-shot adaptation for rare word recognition in end-to-end ASR systems~\cite{lux2021metalearning}. 

Integrating Large Language Models (LLMs) into ASR systems presents a promising avenue to enhance rare word recognition. LLMs, with their extensive training on vast textual data, have a rich understanding of language nuances, including rare words and contextual relationships~\cite{yang2024ctcassisted}. By leveraging LLMs, ASR systems can potentially improve transcription accuracy, benefiting downstream tasks such as keyword spotting, intent detection, and text summarization.

This study investigates the integration of LLMs with ASR systems to enhance rare word recognition. We used a large-scale dataset comprising 190,000 hours of diverse speech data to train and evaluate our models. Our approach involves combining a speech encoder with an LLM decoder within an encoder-decoder architecture. Through comprehensive experiments and ablation studies, we aim to:
\begin{itemize}
    \item Demonstrate that the integration of LLM with ASR systems improves the Rare Word Error Rate (RWER) for rare words, on the scale of a large training dataset.
    \item Figure out the different roles of the speech encoder and LLM in determining overall WER performance and rare word recognition performance.
    \item Present how the improved data quality contributes to the rare word speech recognition.
\end{itemize}

Our findings contribute to the development of more robust LLM-based ASR systems capable of accurately recognizing rare words, thereby enhancing the utility of ASR in various applications.

\section{Related Work}
\label{sec:related}

Recognizing rare words in Automatic Speech Recognition (ASR) systems remains a significant challenge due to their infrequent occurrence in training data. Various strategies have been proposed to address this issue, including post-processing methods, meta-learning approaches, and the integration of Large Language Models (LLMs).

Post-processing techniques focus on error detection and context-aware correction to enhance rare word recognition. For example, He et al. introduced an ASR post-processing method that targets predicted error positions and leverages a rare word list to provide additional contextual knowledge, resulting in improved recognition of rare words~\cite{he2023edcec}.

Meta-learning approaches have also been explored to enable few-shot adaptation for rare word recognition in end-to-end ASR systems. Lux and Vu proposed a method that generates meaningful embeddings for speech and adapts meta-learning algorithms to perform keyword spotting in continuous signals, thus improving the recognition of rare words~\cite{lux2021metalearning}.

The integration of LLMs into ASR systems has gained attention as a means to enhance transcription accuracy, particularly for rare words. For instance, Pu et al. \cite{pu2023multistage} proposed an approach that utilizes the reasoning capabilities of LLMs to improve transcription accuracy. Yang et al. proposed a CTC-assisted LLM-based contextual ASR model that uses coarse CTC decoding results to filter potential relevant hotwords and incorporate them into LLM prompt input, demonstrating significant improvements in recognizing rare long-tail words~\cite{yang2024ctcassisted}. Additionally, Min and Wang explored the potential of using LLMs' in-context learning capabilities to enhance ASR performance, though their findings indicated challenges in leveraging LLMs for error correction in speech recognition transcriptions~\cite{min2023exploring}.

Despite these advancements, challenges persist in effectively integrating LLMs with ASR systems to improve rare word recognition. This study aims to build upon existing research by investigating the roles of speech encoders and LLMs in determining overall Word Error Rate (WER) performance and rare word recognition, utilizing a large-scale dataset to provide empirical evidence for the efficacy of LLM-ASR integration.

\section{Methodology}
\label{sec:method}

We examined a common LLM-ASR setting, drawing inspiration from the SLAM-ASR~\cite{ma2024slam_asr} framework , with an additional step using LoRA for continued fine-tuning after adapter training. For comparison, we utilized a Zipformer Transducer model~\cite{zipformer} , trained on the same 80-dimensional filter bank (fbank) features.

\subsection{Comparison Model}
The Zipformer Transducer model was selected for comparison due to its status as a state-of-the-art architecture, trained from scratch. It was fully trained over approximately 50 epochs, achieving convergence. The model’s configuration aligns with the “large” specification in k2 ~\cite{k2}.

\subsection{Model Architecture}
The model architecture is an encoder-decoder structure with an intermediary adapter module. The architecture consists of the following key components:

\begin{itemize}
    \item \textbf{Encoder}: We employ Whisper V2 as the speech encoder, which processes raw speech FBank features into high-dimensional representations. Whisper V2 has demonstrated robust performance in various ASR tasks and serves as a strong foundation for extracting acoustic features. We opted for Whisper V2 over V3 because our features were pre-calculated, leading to a dimension mismatch when paired with the Whisper V3 model.

    \item \textbf{Adapter}: Positioned between the speech encoder and the decoder, the adapter module facilitates the alignment of speech encoder features with the input expectations of the language model. It is implemented as a simple multi-layer perceptron (MLP) with activation functions. Additionally, we merged every two frames from the audio encoder into a single frame through the adapter to reduce the audio sequence length, improving computational efficiency.

    \item \textbf{Decoder}: Qwen-7B-Chat~\cite{bai2023qwen} serves as the language model (decoder). It accepts both text instructions and dense speech features generated by the audio encoder. This design allows the decoder to better contextualize audio features within textual prompts.
\end{itemize}

\subsection{Contribution Ablation Studies}
To assess the contributions of each component, we performed the following ablation studies:

\begin{itemize}
    \item \textbf{Baseline Whisper V2}: Evaluated the performance of the original, un fine-tuned Whisper V2 model.
    
    \item \textbf{Fine-tuned Whisper V2 (with original decoder)}: Compared the performance of the fine-tuned Whisper V2 model after completing the first training stage, with the combination of this fine-tuned Whisper V2 encoder and LLM.

    \item \textbf{Further Fine-tuned Whisper V2}: Further fine-tuned Whisper V2 over three epochs and compared its performance with the LLM-ASR model. This comparison allowed us to evaluate the contribution of the LLM under similar training settings.
\end{itemize}

\section{Experimental Setup}
\label{sec:method}

\subsection{Training Dataset}
We utilized a 190,000-hour speech dataset, predominantly sourced from YouTube, encompassing a wide range of speakers, accents, and recording conditions. To enhance transcription accuracy, we applied Whisper V3 for pseudo-labeling, ensuring that each audio segment was accompanied by corresponding text. The dataset underwent rigorous filtering processes, including:

\begin{itemize}
    \item \textbf{WER Filtering}: Calculated Word Error Rate (WER) between whipser-generated labels and video-generated labels, discarding segments with high WER to maintain quality.
    \item \textbf{Simple Sentence Filtering}: Removed overly simplistic sentences to ensure linguistic diversity.
    \item \textbf{Duration Filtering}: Excluded audio segments exceeding 20 seconds to maintain uniformity in training samples since we identified Whisper tends to be over-confident in a longer input.
\end{itemize}

\subsection{Evaluation Dataset}

In this study, we evaluate our model on two datasets: \textbf{Primock57} and \textbf{Kincaid46}. Both datasets include audio recordings, corresponding transcriptions, and rare word annotations manually extracted for evaluation.

\textbf{Primock57:}  
The Primock57 dataset consists of 57 simulated primary care consultations, including audio recordings, manual utterance-level transcriptions, and associated consultation notes~\cite{babylonhealth2023primock57}. In our study, we manually extracted rare words from this dataset, creating two distinct subsets (\textbf{Set 1} and \textbf{Set 2}) to evaluate the models' capabilities in recognizing rare or low-frequency words effectively.

\textbf{Kincaid46:}  
The Kincaid46 dataset comprises 46 audio recordings and their corresponding transcripts ~\cite{kincaid2018transcription}. The dataset covers diverse recording conditions, including scripted and unscripted broadcasts, telephone and VoIP calls, and meetings. Similar to Primock57, we manually extracted rare words from this dataset to assess the models' ability to handle uncommon terminology effectively.

\subsection{Training Procedure}
The training process was executed in three distinct stages:

\begin{enumerate}
    \item \textbf{Speech Encoder Fine-tuning}: Conducted one epoch of fine-tuning on the speech encoder prior to integrating the LLM. This step aimed to mitigate domain adaptation challenges between features and annotations.
    
    \item \textbf{Adapter Training}: Trained the adapter module for one epoch, ensuring effective interfacing between the encoder and decoder.

    \item \textbf{LoRA Fine-tuning}: After adapter training, we introduced a Low-Rank Adaptation (LoRA)~\cite{hu2021lora} phase. During this stage, we fine-tuned the LLM weights for an additional epoch while re-enabling adapter weight tuning. This approach demonstrated improved performance compared to the original SLAM-LLM training strategies with only adapter training enabled.
\end{enumerate}

Learning rates were set at $1 \times 10^{-4}$ for the first two stages and reduced to $1 \times 10^{-6}$ during the LoRA fine-tuning phase.

\subsection{Decoding Strategies}
During inference, we employed beam search decoding with a beam size of 8. To prevent repetitive outputs, an n-gram repetition penalty was applied, configured with $n=6$.

\subsection{Hardware}
Experiments were conducted on a system equipped with 8 NVIDIA H100 GPUs, utilizing DeepSpeed Stage 2 optimization~\cite{rasley2020deepspeed} to manage memory efficiency and computational speed.

\noindent
\FloatBarrier

\section{Results and Analysis}
\label{sec:results}

In this section, we present the experimental results on two datasets: \textbf{Primock57} and \textbf{Kincaid46}. We focus on three key evaluation metrics: Orthographic Word Error Rate (\textbf{O-WER}), Normalized Word Error Rate (\textbf{N-WER}), and Rare Word Error Rate (\textbf{R-WER}). Additionally, we analyze the contributions of the speech encoder, the language model (LLM), and fine-tuning strategies through comprehensive ablation studies.

\subsection{Comparison Between LLM-ASR and Transducer-Zipformer}

We compare the LLM-ASR architecture with the Transducer-Zipformer architecture on two datasets. The results reveal significant differences in their behavior across various metrics.

\begin{table*}[t!] 
\centering
\caption{Performance Comparison Between Zipformer-Transducer and LLM-ASR on Primock57 and Kincaid46}
\label{tab:combined_results}
\resizebox{0.6\linewidth}{!}{
\begin{tabular}{llcc}
\toprule
\textbf{Dataset} & \textbf{Metric} & \textbf{Zipformer-Transducer} & \textbf{LLM-ASR} \\
\midrule
\multirow{4}{*}{\textbf{Primock57}} 
    & O-WER (\%) & 26.7 & \textbf{25.3} \\
    & N-WER (\%) & \textbf{6.7} & 7.7 \\
    & R-WER Set 1 (\%) & 7.3 & \textbf{6.8} \\
    & R-WER Set 2 (\%) & 40.2 & \textbf{32.2} \\
\midrule
\multirow{3}{*}{\textbf{Kincaid46}} 
    & O-WER (\%) & \textbf{17.4} & 18.0 \\
    & N-WER (\%) & \textbf{9.1} & 9.3 \\
    & R-WER (\%) & 18.7 & \textbf{17.0} \\
\bottomrule
\end{tabular}
}
\end{table*}

\begin{table*}[t!]
\centering
\caption{Ablation Study Results Across Primock57 and Kincaid46 Datasets}
\label{tab:ablation_results}
\resizebox{\linewidth}{!}{
\begin{tabular}{llcccc}
\toprule
\textbf{Dataset} & \textbf{Metric} & \textbf{Baseline Whisper V2} & \textbf{Encoder Fine-tuned (1 Epoch)} & \textbf{Encoder Fine-tuned (3 Epochs)} & \textbf{LLM-ASR} \\
\midrule
\multirow{4}{*}{\textbf{Primock57}} 
    & \textbf{O-WER (\%)} & 28.1 & 25.7 & 25.4 & \textbf{25.3} \\
    & \textbf{N-WER (\%)} & 9.2 & 7.7 & 7.7 & 7.7 \\
    & \textbf{R-WER Set 1 (\%)} & 8.8 & 7.4 & 7.2 & \textbf{6.8} \\
    & \textbf{R-WER Set 2 (\%)} & 36.8 & 35.6 & 35.6 & \textbf{32.2} \\
\midrule
\multirow{3}{*}{\textbf{Kincaid46}} 
    & \textbf{O-WER (\%)} & 20.0 & \textbf{17.8} & 18.0 & 18.0 \\
    & \textbf{N-WER (\%)} & 10.9 & 9.2 & 9.2 & 9.3 \\
    & \textbf{R-WER (\%)} & 21.5 & 18.6 & 18.0 & \textbf{17.0} \\
\midrule
\end{tabular}
}
\end{table*}

\textbf{Performance on General Metrics (O-WER and N-WER)}  
Both architectures perform comparably on general error metrics, including O-WER and N-WER, across both datasets. On \textbf{Primock57}, LLM-ASR achieves a slight reduction in O-WER (25.3\% vs. 26.7\%) but experiences a slight increase in N-WER (7.7\% vs. 6.7\%). On \textbf{Kincaid46}, the differences in O-WER (18.0\% vs. 17.4\%) and N-WER (9.3\% vs. 9.1\%) are marginal.

These results indicate that LLM-ASR, while competitive with Transducer-Zipformer on common transcription tasks, does not inherently outperform it in these general metrics.

\textbf{Performance on Rare Words (R-WER)}  
LLM-ASR shows a significant advantage in \textbf{R-WER}. On \textbf{Primock57}, in Set 1 achieves a reduction from 7.3\% to 6.8\%, while Set 2 sees a dramatic improvement from 40.2\% to 32.2\%. Similar trends are observed in \textbf{Kincaid46}, where R-WER decreases from 18.7\% to 17.0\%.

This improvement suggests that the integration of the LLM introduces better contextual reasoning and semantic understanding, enabling the system to recognize and transcribe rare or out-of-vocabulary words more effectively. In contrast, Transducer-Zipformer seems limited in its ability to handle such cases, relying more on acoustic information than linguistic context.

In conclusion, LLM-ASR demonstrates a clear advantage in zero-shot rare word recognition tasks, particularly in challenging datasets like \textbf{Primock57 Set 2}. On general transcription metrics (\textbf{O-WER}, \textbf{N-WER}), both architectures perform similarly, emphasizing that LLM is not that helpful when discussing general transcription metrics.

\subsection{Ablation Study Analysis}

To better understand the contributions of each component in LLM-ASR architecture, as previously mentioned, we conducted ablation studies by analyzing multiple training strategies.

\textbf{Fine-tuned whisper encoder and decoder}  
Compared with the baseline whisper model, the fine-tuned model leads to significant improvements in \textbf{O-WER} and \textbf{N-WER}, especially in the early epochs. On \textbf{Primock57}, the O-WER decreases from 28.1\% to 25.7\% after one epoch of fine-tuning. However, additional epochs result in diminishing returns, with only marginal improvements observed between one epoch (25.7\%) and three epochs (25.4\%).

In contrast, improvements in \textbf{R-WER} from encoder fine-tuning are limited. This indicates that while encoder fine-tuning refines the acoustic feature representation and overall transcription quality, it lacks the ability to significantly improve rare word recognition. This suggests that rare word recognition relies more heavily on linguistic reasoning, a task better suited for the LLM.

\textbf{LLM integration}  
This stage brings noticeable improvements in \textbf{R-WER}, reducing it to 6.7\% on \textbf{Primock57 Set 1} and 32.2\% on \textbf{Primock57 Set 2}, and 17.0\% on \textbf{Kincaid46}. These improvements highlight the effectiveness of LoRA in adapting LLMs on the transcription task.

To summarize it, speech encoder primarily determines the overall transcription performance (\textbf{O-WER}, \textbf{N-WER}). The adapter facilitates feature alignment and improves rare word recognition. LLM provides impactful improvements in rare word recognition (\textbf{R-WER}).

\subsection{Impact of Data Quality}

We also observed the influence of data quality by comparing the baseline Whisper V2 model with fine-tuned versions. The baseline model, despite being pre-trained on diverse datasets, performs poorly on all metrics when applied directly to the target datasets. Fine-tuning the encoder, adapter, and LLM successively brings significant improvements.

\subsection{Insights and Future Work}

Our analysis highlights several key insights. LLM-ASR excels in rare word recognition, demonstrating significant improvements in zero-shot scenarios, while the speech encoder remains the primary driver of overall transcription quality. High-quality labeled data emerges as an essential factor, reinforcing the need for meticulous data preparation in large-scale ASR tasks.  

Looking forward, our future work will focus on optimizing LLM alignment for rare word recognition, reducing decoding latency while maintaining transcription accuracy, and exploring domain adaptation techniques to enhance model robustness across diverse datasets and acoustic conditions.

\bibliographystyle{IEEEbib}
\bibliography{main}

\end{document}